\documentclass[10pt,twocolumn,letterpaper]{article}

\usepackage{cvpr}      

\usepackage{graphicx}
\usepackage{amsmath}
\usepackage{amssymb}
\usepackage{booktabs}
\usepackage{bm}

\usepackage{cuted}
\usepackage{capt-of}

\usepackage[pagebackref,breaklinks,colorlinks]{hyperref}

\usepackage[capitalize]{cleveref}
\crefname{section}{Sec.}{Secs.}
\Crefname{section}{Section}{Sections}
\Crefname{table}{Table}{Tables}
\crefname{table}{Tab.}{Tabs.}

\def\cvprPaperID{3239} 
\def\confName{CVPR}
\def\confYear{2023}

\usepackage{overpic}
\usepackage{enumitem} 
\usepackage{overpic} 
\usepackage{color}
\usepackage{xcolor}
\usepackage{array}
\usepackage{bm}

\newcommand{\ignorethis } [1] {}

\definecolor{turquoise}{cmyk}{0.65,0,0.1,0.3}
\definecolor{purple}{rgb}{0.65,0,0.65}
\definecolor{dark_green}{rgb}{0, 0.5, 0}
\definecolor{orange}{rgb}{0.8, 0.6, 0.2}
\definecolor{red}{rgb}{0.8, 0.2, 0.2}
\definecolor{darkred}{rgb}{0.6, 0.1, 0.05}
\definecolor{blueish}{rgb}{0.0, 0.3, .6}
\definecolor{light_gray}{rgb}{0.7, 0.7, .7}
\definecolor{pink}{rgb}{1, 0, 1}
\definecolor{greyblue}{rgb}{0.25, 0.25, 1}

\definecolor{color1}{rgb}{0.36470588235, 0.7294117647, 0.43921568627}
\definecolor{color2}{rgb}{0.88235294117, 0.27450980392, 0.2431372549}
\definecolor{color3}{rgb}{0.58823529411, 0.46666666666, 0.4}
\definecolor{color4}{rgb}{0.30196078431, 0.62352941176, 0.88235294117}
\definecolor{color5}{rgb}{0.8862745098, 0.66274509803, 0.22745098039}
\newcommand{\PaletteNeRF}{{\color{color1}P\color{color2}a\color{color3}l\color{color4}e\color{color5}t\color{color5}t\color{color4}e}NeRF}

\newcommand{\comment}[1]{}

\begin{document}

\title{\PaletteNeRF: Palette-based Appearance Editing of Neural Radiance Fields}


\author{
Zhengfei Kuang\textsuperscript{1}\thanks{Parts of this work were done when Zhengfei Kuang was an intern at Adobe Research.}, Fujun Luan\textsuperscript{2}, Sai Bi\textsuperscript{2}, Zhixin Shu\textsuperscript{2}, Gordon Wetzstein\textsuperscript{1}, Kalyan Sunkavalli\textsuperscript{2}\\
\textsuperscript{1}Stanford University\qquad 
\textsuperscript{2}Adobe Research\\
{\tt\small \{zhengfei,gordonwz\}@stanford.edu \qquad \{fluan,sbi,zshu,sunkaval\}@adobe.com}\\
\\
{\tt\small \url{https://palettenerf.github.io}}
}


\maketitle


\begin{strip}\centering
\vspace{-50px}
\includegraphics[width=\textwidth]{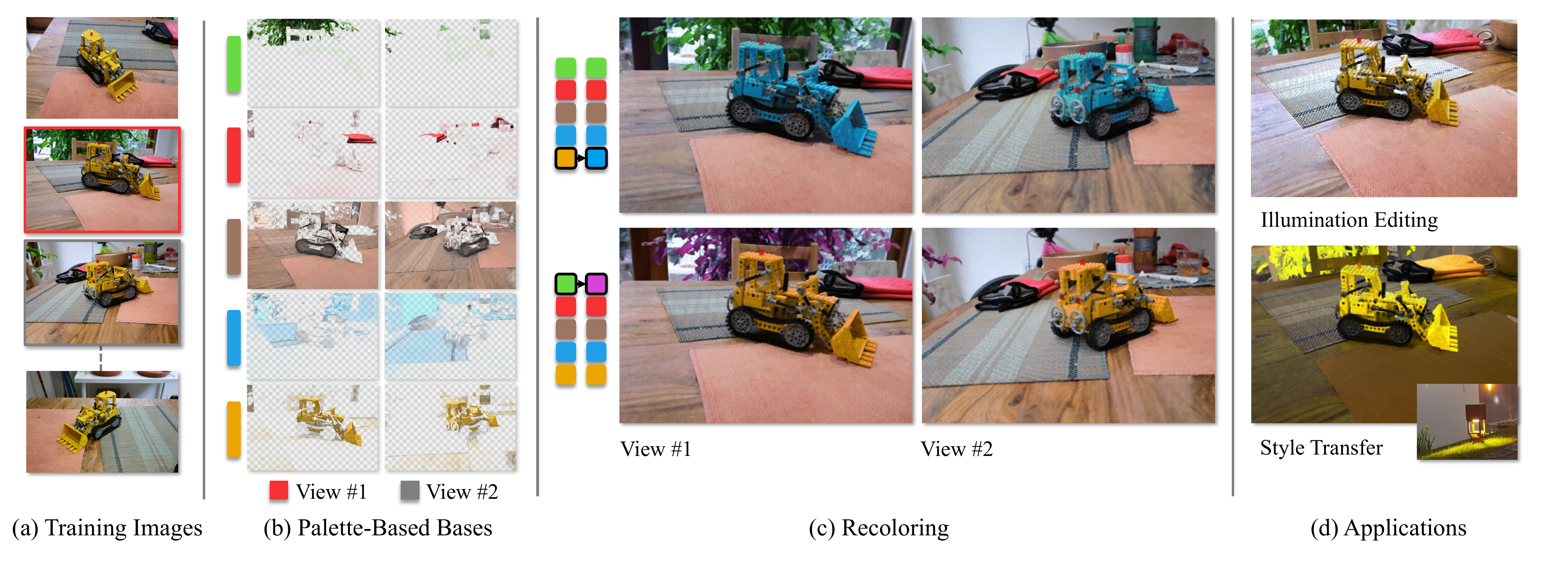}
\vspace{-10px}
\captionof{figure}{We propose \PaletteNeRF, a novel method for efficient appearance editing of neural radiance fields (NeRF). Taking (a) multi-view photos as training input, our approach reconstructs a NeRF and decomposes its appearance into a set of (b) 3D palette-based color bases. This enables (c) intuitive and photorealistic recoloring of the scene with 3D consistency across arbitrary views. Further, we show that (d) our method supports various palette-based editing applications such as illumination modification and 3D photorealistic style transfer.  
\label{fig:teaser}}
\end{strip}

\begin{abstract}
Recent advances in neural radiance fields have enabled the high-fidelity 3D reconstruction of complex scenes for novel view synthesis. However, it remains underexplored how the appearance of such representations can be efficiently edited while maintaining photorealism. 
In this work, we present PaletteNeRF, a novel method for photorealistic appearance 
editing of neural radiance fields (NeRF) based on 3D color decomposition. 
Our method decomposes the appearance of each 3D point into a linear
combination of palette-based bases (i.e., 3D segmentations defined by a group of NeRF-type functions) that are shared across the scene. While our palette-based bases are view-independent, we also predict a view-dependent function to capture the color residual (e.g., specular shading). During training, we jointly optimize the basis functions and the color palettes, and we also introduce novel regularizers to encourage the spatial coherence of the decomposition. Our method allows users to efficiently edit the appearance of the 3D scene by modifying the color
palettes. We also extend our framework with compressed semantic features for 
semantic-aware appearance editing. We demonstrate that our technique is superior to 
baseline methods both quantitatively and qualitatively for appearance editing 
of complex real-world scenes. 



\end{abstract}
\section{Introduction}

Neural Radiance Fields (NeRF)~\cite{NeRF} and its variants
~\cite{PlenOxels, Instant-NGP, KiloNeRF, TensoRF} have received
increasing attention in recent years for their ability to robustly reconstruct real-world 3D scenes from 2D images and enable high-quality, photorealistic novel view synthesis.
However, such volumetric representations are challenging to edit due to the fact that scene appearance is implicitly encoded in neural features and network weights that do not support local manipulation or intuitive modification.  

Multiple approaches have been proposed to support editing of NeRF. One category 
of methods~\cite{NeRFactor,NeROIC,NeRD,PhySG} recover the material properties of the scene so that they can re-render them under novel lighting conditions or adjust the material properties
such as surface roughness. Such methods rely on accurate estimation of the 
scene reflectance, which is typically challenging for 
real-world complex scenes captured under unconstrained environment. Another category of
methods~\cite{CLIP-NeRF,NeRF-W} learns a latent code on which NeRF can be conditioned to produce the desired  
appearance. However, these methods often suffer from limited capacity and flexibility and do 
not support fine-grained editing. In addition, some other methods~\cite{ARF} learn to transfer the appearance of NeRF to match a given style image, but sometimes
fail to maintain the same level of photorealism in the original scene.

In this paper, we propose \emph{\PaletteNeRF}, a novel method to 
support flexible and intuitive editing of NeRF. Our method is inspired by 
previous image-editing methods based on color 
palettes~\cite{chang2015palette, RGBXY}, where a small 
set of colors are used to represent the full range of colors in the image.  
We model the radiance of each point using a combination 
of specular and diffuse components, and we further decompose the
diffuse component into a linear combination of view-independent color bases 
that are shared across the scene.
During training, we jointly optimize the per-point specular component, the global color bases and the per-point linear weights to minimize the difference between the rendered images and the ground truth images. We also introduce novel regularizers 
on the weights to encourage the sparseness and spatially coherence of the decomposition and 
achieve more meaningful grouping. With the proposed framework, we can intuitively 
edit the appearance of NeRF by freely modifying the learned color 
bases (Fig.~\ref{fig:teaser}). We further show that our framework can be combined with 
semantic features to support semantic-aware editing. Unlike previous palette-based image~\cite{RGBXY, FSCS} or video~\cite{du2021video} editing methods, our method produces more globally coherent and 3D consistent recoloring results of the scene across arbitrary views. We demonstrate that our method can enable more fine-grained local color 
editing while faithfully maintaining the photorealism of the 3D scene, and 
achieves better performance than baseline methods both quantitatively and qualitatively. In summary, our
contributions include:

\begin{itemize}
\item We propose a novel framework to facilitate the editing of NeRF by decomposing the radiance field into a weighted combination of learned color bases.

\item We introduced a robust optimization scheme with novel regularizers to achieve intuitive decompositions.

\item Our approach enables practical palette-based appearance editing, making it possible for novice users to interactively edit NeRF in an intuitive and controllable manner on commodity hardware.


\end{itemize}

\section{Related Work}


\paragraph{Neural radiance fields.}
Neural radiance fields in the form of MLPs have been extensively used for neural 
rendering tasks such as novel view synthesis.  Typically, these methods 
\cite{VolSDF, NeuS, NeRF,sitzmann2019srns} encode the geometry and appearance 
of the scene into network weights of the MLPs. Many recent works 
~\cite{PlenOxels, TensoRF, Point-NeRF, zhang2022nerfusion, Instant-NGP} propose
to speed up the training and improve the performance of the models by applying 
a combination of light-weight MLPs and neural feature maps or volumes. 
However, different from traditional graphics primitives such as triangle meshes,
both the neural features and the network weights represent scene appearance
in an implicit manner and do not support intuitive editing or controls such as recoloring, thereby greatly limiting their applications in existing graphics pipelines.

\paragraph{Appearance editing with NeRF.} 
Many methods have been proposed to support appearance editing of NeRF. 
Some methods~\cite{NeRFactor, NeRV, NeRD, Neural-PIL} recover the physical 
properties of the scene such as albedo, specular roughness, and then they 
can support rendering the scene under novel lighting conditions or changing  
its material properties. Some other works~\cite{NeRF-W, CLIP-NeRF, AE-NeRF}
learn a latent code jointly with the NeRF reconstruction so as to control 
its appearance such as changing the illuminations or colors by taking
new latent codes as input, which can be mapped from user input or interpolated 
from existing latent codes. Moreover, there are also approaches~\cite{ARF, StyleNeRF, UPST-NeRF} that try to 
optimize the NeRF to match its appearance against the provided 
style images. All these methods do not support fine-grained intuitive color 
editing of NeRF as ours. Concurrently, Ye et al.~\cite{ye2022intrinsicnerf} introduces a NeRF-based intrinsic decomposition model which enables 3D intuitive recoloring, but it does not support palette-based editing.


\paragraph{Palette-based editing.}
Color palette-based 
methods~\cite{chang2015palette,DBLP:conf/siggrapha/FurusawaHOO17,DBLP:journals/tip/ZhangXST17,DBLP:journals/tog/ZouMGDF19,DBLP:journals/tog/TanLG17,RGBXY}
have been previously used in 2D image editing tasks such as recoloring. 
However, it is non-trivial to  adapt such methods for NeRF editing and  
naively performing the editing on each rendered frame cannot guarantee 
view-consistent results. Instead, we integrate the learning of the color bases 
and the spatially varying weights into the NeRF optimization process, 
thereby enabling view-consistent and 3D-aware editing. A recent work~\cite{DBLP:journals/cgf/TojoU22} proposes a palette-based recoloring method on NeRF with posterization. However, it is limited to non-photorealistic editing only.

\section{Method}

\begin{figure*}
\centering
\includegraphics[width=1.0\linewidth]{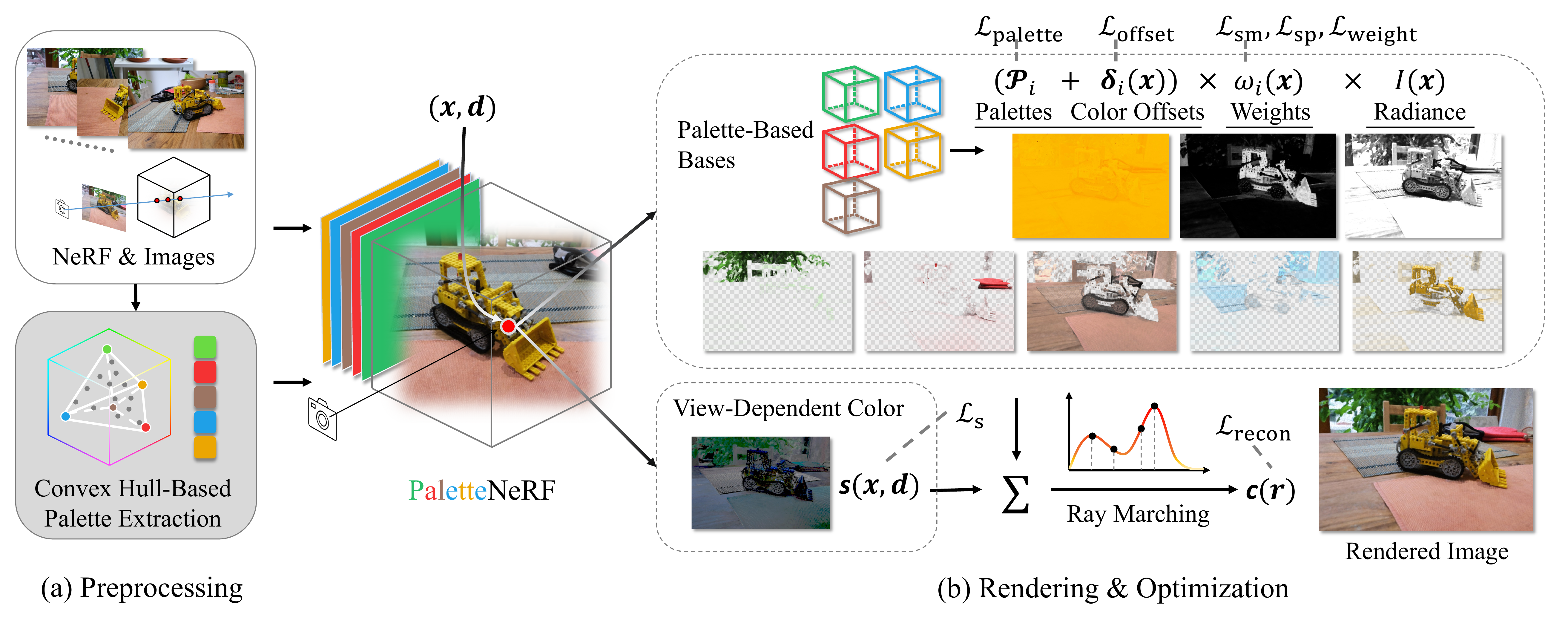}
\vspace{-20px}
\caption{\textbf{The overview of our pipeline}. Given a set of training images, we first (a) reconstruct the scene geometry and build the color palettes with existing methods. Then, our \PaletteNeRF{} (b) decomposes the scene appearance into multiple palette-based bases and the view-dependent color. We deploy a series of losses on the palette-based base's functions, the view-dependent color, and the final output. }
\label{fig:overview}
\end{figure*}

Fig.~\ref{fig:overview} illustrates the overview of our multi-stage pipeline.  Given a set of images with known poses from a scene, we first optimize a NeRF-based model to reconstruct the geometry of the scene. Then, we extract $N_p$ color palettes with the input images and the learned scene geometry. Finally, we train a segmentation model to decompose the scene appearance into multiple bases based on the extracted palettes. Our decomposition result is able to drive various downstream applications, such as recoloring, photorealistic style transfer and illuminance modification. 

\subsection{Volumetric Rendering}
We build our model on the widely used framework Neural Radiance Fields (NeRF). Typically, a NeRF-based model optimizes two neural functions: a geometry function $\sigma(\bm{x})$ and a color function $\bm{c}(\bm{x}, \bm{d})$. The geometry function takes a 3D position as input and outputs the density at this point. The color function takes a 3D position and a viewing direction as input and output a corresponding RGB color.
To render an image from a given camera pose, NeRF samples batches of rays from the camera to render the pixels. For each sampled ray $\bm{r} = (\bm{o}, \bm{d})$, a group of 3D points are sampled along the ray path $\bm{x}_{1..M}$ with depth $t_{1..M}$, where $\bm{x}_i = \bm{o} + t_i \bm{d}$. The color prediction of the ray $\bm{r}$ is calculated as:
\begin{equation}
    \bm{c}(\bm{r})=\sum_{i=1}^{M}\alpha_i(1-w_i)\bm{c}(\bm{x}_i, \bm{d}),
    \label{eq:NeRF}
\end{equation}
where 
$w_i=\exp(-(t_i-t_{i-1})\sigma(\bm{x}_i))$ is the transmittance of the ray between the $i$'th sample point and the $(i+1)$'th sample point, and $\alpha_i=\prod_{j=1}^{i-1}w_i$ is the ray attenuation from the ray's origin to the $i$'th sample point. 

\begin{figure}
\centering
\includegraphics[width=0.495\textwidth]{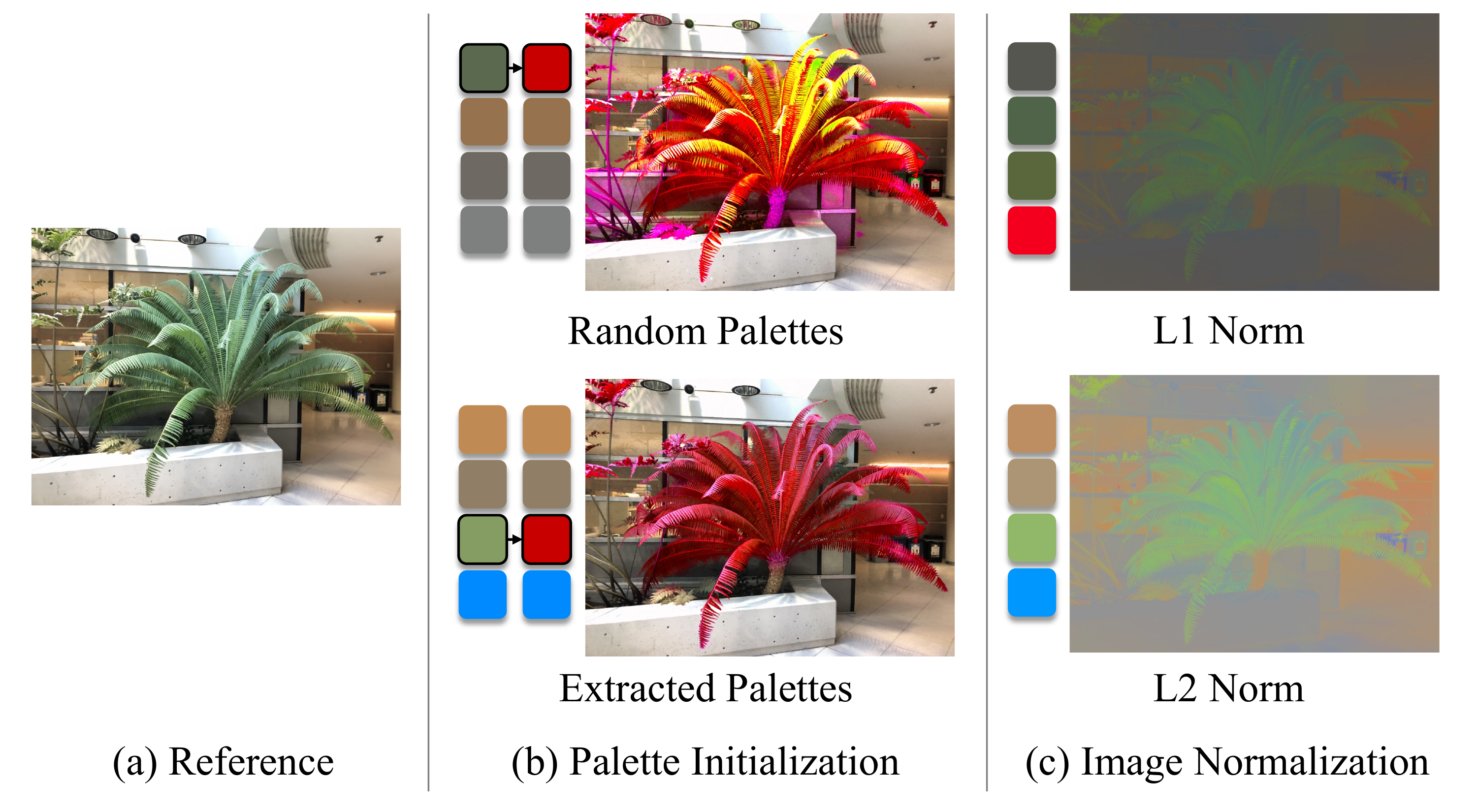}
\caption{\textit{Left:} Reference image; \textit{Middle:} Comparison between the recoloring result of our model trained with extracted palettes and model trained with random initialized palettes; \textit{Right:} Normalized training image and the extracted palettes from normalization of different levels.}
\label{fig:palette_extraction}
\end{figure}

\subsection{Palette Extraction}
Although it is possible to directly train our model with randomly initialized color palettes, doing this will bring excessive ambiguity to the problem and may produces irregular results as shown in Fig.~\ref{fig:palette_extraction}(b). Thankfully, the problem of extracting color palettes from images has been extensively researched over the past years, thus we use the extraction method from a state-of-the-art image recoloring work~\cite{RGBXY} as our initialization. In general, this method extracts color palettes from the 3D convex hull of the clustered image colors in the RGB space.
In our scenario, we simply select all training image pixels with valid depth from NeRF's depth maps and concatenate their color as input.

We notice that with input from multiple images, the extraction method may produce palettes that are chromatically similar (e.g., palettes including a light yellow and a dark yellow), due to the varying shading of the captured scene. This may result in unrealistic appearance editing. Inspired by image illumination decomposition works~\cite{carroll2011illumination, Meka:2021}, we normalize the input colors of the training images by their intensity, to narrow down the search space of color palettes. Conventionally, the intensity of a color is represented by the L1 norm of its RGB value. However, normalizing with L1 norm projects the colors to a plane, which is a highly ill-posed edge case for 3D convex hull calculation. These problems can be addressed by replacing the norm with a higher order one. Empirically, we find that palettes extracted from L2 normalized images work well in our next decomposition stage, and use them in all of our experiments.

In addition to the extracted palettes $\bar{\bm{\mathcal{P}}}$, we also keep the blending weights $\bar\omega$ of the input pixel colors, calculated from the method of the same work. These weights act as an additional supervision in the next stage.

\subsection{Color Decomposition}

\begin{figure}
\centering
\includegraphics[width=\linewidth]{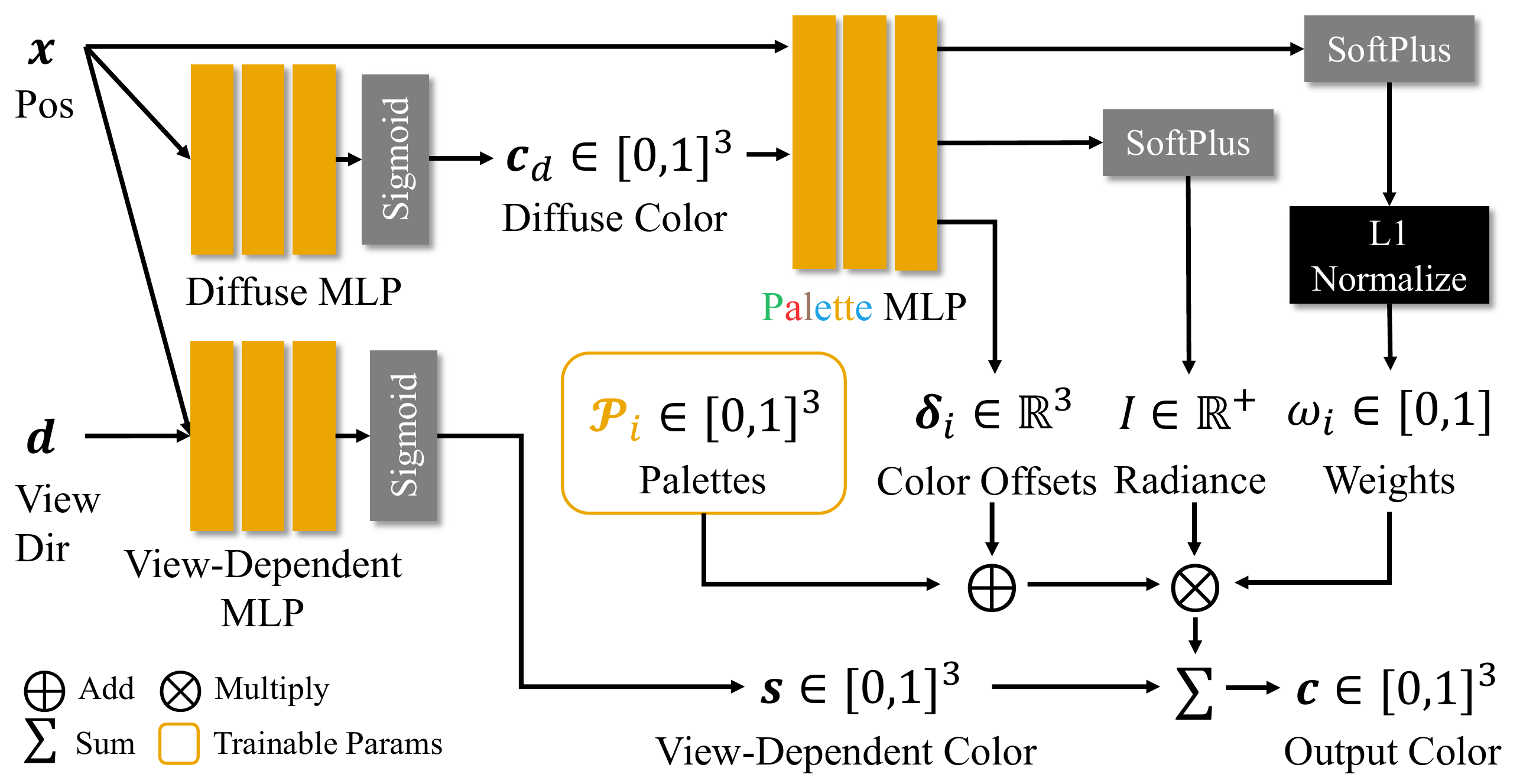}
\caption{\textbf{Our network structure.} As we only show one palette-based basis in the figure, our network generates $N_p$ bases in parallel and they are summed up in the final step.}
\label{fig:basis_segmentation}
\end{figure}

To introduce our color decomposition model, we first explain our target outputs, then describe the structure of our decomposition network. 

As shown in Fig.~\ref{fig:overview}, given the color palettes of size $N_p$, our model aims to reconstruct $N_p$ view-independent palette-based bases and an additional view-dependent color function representing all view-dependent shading such as specular reflections. The palette-based bases correspond to the extracted color palettes, and are defined by two functions of $\bm{x}$: A color offset function $\bm{\delta}: \mathbb{R}^3 \rightarrow \mathbb{R}^3$, and a weight function $\omega: \mathbb{R}^3 \rightarrow [0,1]$. Observing that realistic image captures usually consists of a huge variety of colors, we allow the basis color of each point shift from the palette color with an offset (inspired by image-based soft color segmentation methods ~\cite{aksoy2017unmixing, FSCS}). This design increases the capacity of the bases and benefits the segmentation quality on complex scenes. We also introduce an intensity function $I: \mathbb{R}^3 \rightarrow [0,1]$ that is shared among all palette-based bases, due to the normalization of the extracted palettes. While the aforementioned functions only take position as input, our model also contains a view-dependent color function $\bm{s}: \mathbb{R}^5 \rightarrow [0,1]^3$ that takes viewing direction as input, too. We compose all of these bases into a color output $\bm{c}\in \mathbb{R}^3$ by the following equation:
\begin{equation}
\bm{c}(\bm{x}, \bm{d}) = \bm{s}(\bm{x}, \bm{d}) + I(\bm{x}) \sum_{i=1}^{N_p} \omega_i(\bm{x}) \left(\bm{\mathcal{P}}_i + \bm{\delta}_i(\bm{x})\right),
\label{eq:color_segmentation}
\end{equation}
where $\omega_i(\bm{x})$ are normalized by their sum. We clamp the summed-up color $\bm{c}$ to $[0, 1]$ as the final output. We also optimize the palette color $\bm{\mathcal{P}}_i$ during the training. As shown in Fig.~\ref{fig:basis_segmentation}, Our network consists of three MLPs: the Diffuse MLP predicts the diffuse color $\bm{c}_d(\bm{x})$, i.e. the sum of all palette-based bases; the View-Dependent MLP generates view-dependent color $\bm{s}(\bm{x}, \bm{d})$; Finally, the Palette MLP predicts the function values of the palette-based bases: $\omega_i(\bm{x}), \bm{\delta_i}(\bm{x})$ and $I(\bm{x})$, where $\bm{c}_d$ is also fed as an input prior. We compose all of the network outputs to the final color using Eq.~\ref{eq:color_segmentation}.

\subsection{Optimization}

Given the fact that separating multiple bases from the scene appearance is a considerably ill-posed task, a lot of cautions should be taken in designing the optimization scheme. Hence, we develop a series of losses to regulate the optimized parameters and to avoid undesirable results such as local-minimum. 

To begin with, we deploy the image reconstruction loss defined as:
\begin{equation}
\mathcal{L}_{
\textrm{recon}} = \lVert \bm{c}^{\textrm{ref}} - \bm{c}(\bm{r})\rVert _2^2 + \lVert \bm{c}^{\textrm{ref}} - \left(\bm{c}_d(\bm{r})+\bm{s}(\bm{r})\right)\rVert _2^2,
\label{eq:img_loss}
\end{equation}
where $\bm{c}^{\textrm{ref}}$ is the ground truth color, $\bm{c}(\bm{r}), \bm{c}_d(\bm{r}), \bm{s}(\bm{r})$ are calculated from the volume rendering equation at Eq.~\ref{eq:NeRF}. The second term of this loss can also be considered as the L2 distance between $\bm{c}_d(\bm{r})$ and the sum of palette-based bases. 
We also add a regularization loss $\mathcal{L}_\textrm{s}$ to the view-dependent color function $\bm{s}(\bm{x}, \bm{r})$, to prevent the particular situation where $\bm{s}$ dominates the appearance and pushes all palette-based bases to 0. This loss is defined as:
\begin{equation}
\mathcal{L}_\textrm{s} = \lVert \bm{s}(\bm{x}, \bm{d}) \rVert _2^2.
\label{eq:s_loss}
\end{equation}

While our model uses color offsets to shift the basis color, it is necessary to restrict the blending weights and color offset to avoid extreme solutions. Hence, we adapt the sparsity loss $\mathcal{L}_{\textrm{sp}}$ and the color offset loss $\mathcal{L}_{\textrm{offset}}$ from the image soft segmentation method \cite{aksoy2017unmixing} . They are defined as: 
\begin{align}
\mathcal{L}_{\textrm{sp}} &= \frac{\sum_{i=1}^{N_p} \omega_i(\bm{x})}{\sum_{i=1}^{N_p} \omega_i^2(\bm{x})} - 1,
\label{eq:sparse_loss} \\
\mathcal{L}_{\textrm{offset}} &= \lVert \bm{\delta}(\bm{x}) \rVert _2^2.
\label{eq:offset_loss}
\end{align}
The sparsity loss aims to make the blending weights sparser (e.g., segmenting each point $\bm{x}$ to fewer bases), which will eventually increase the capacity of the bases by increasing the color offset. On the other hand, the color offset loss directly suppresses the magnitude of color offsets, preventing them from deviating from the palettes too much. Intuitively, these two losses act as two adversarial roles, and finding a good balance between them will lead to neat segmentation results with reasonable basis color. In our experiments, however, we observe that these two losses may lead to harsh segmentation results, which will drastically affect the quality of the following editing (see our qualitative ablation study). Thus we introduce a novel 3D-aware smooth loss to smooth the weight function based on the NeRF's output. It is given by:
\begin{equation}
\mathcal{L}_{\textrm{sm}} = \xi(\bm{x}, \bm{x}+\bm{\varepsilon}) \lVert \bm{\omega}(\bm{x}) - \bm{\omega}(\bm{x}+\varepsilon) \rVert ^2,
\label{eq:delta_loss}
\end{equation}
where $\bm{\omega}=\omega_{1...N_p}, \bm{\varepsilon}$ is a random position offset sampled from a Gaussian distribution, and $\xi(\cdot)$ is the similarity between two points. Here, we adapt the Gaussian kernel used in the bilateral filter, and define the similarity function as:
\begin{equation}
\xi(\bm{x}, \bm{y}) = \exp(-\frac{\lVert\bm{x} - \bm{y}\rVert^2}{\sigma_x} - \frac{\lVert \bm{c}_d(\bm{x}) - \bm{c}_d(\bm{y})\rVert^2}{\sigma_c}),
\label{eq:delta_loss}
\end{equation}
where $\sigma_x$ and $\sigma_c$ are smoothing parameters. While the diffuse color $\bm{c}_d$ is used in the smooth loss, we cut off their gradients during the training. 

Finally, we add two more losses that incorporate the supervisions from the palette extraction model. They are:
\begin{align}
\mathcal{L}_{\textrm{palette}} &= \lVert \bm{\mathcal{P}} - \bar{\bm{\mathcal{P}}} \rVert _2^2, \\
\mathcal{L}_{\textrm{weight}} &= \lVert \omega - \bar{\omega} \rVert _2^2.
\label{eq:palette_loss}
\end{align}
As a summary, the overall loss of our optimization is the weighted sum defined as:
\begin{equation}
    \begin{split}
    \mathcal{L} &= \mathcal{L}_\textrm{recon} 
    + \lambda_\textrm{s} \mathcal{L}_\textrm{s} 
    + \lambda_{\textrm{sp}} \mathcal{L}_{\textrm{sp}} \\
    & + \lambda_{\textrm{offset}} \mathcal{L}_{\textrm{offset}} 
    + \lambda_{\textrm{sm}} \mathcal{L}_{\textrm{sm}} \\
    & + \lambda_{\textrm{palette}} \mathcal{L}_{\textrm{palette}}
    + \lambda_{\textrm{weight}} \mathcal{L}_{\textrm{weight}},
    \label{eq:overall_loss}
    \end{split}
\end{equation}
where $\lambda_\textrm{s}, \lambda_{\textrm{sp}}, \lambda_{\textrm{offset}}, \lambda_{\textrm{sm}}, \lambda_{\textrm{palette}}, \lambda_{\textrm{weight}}$ are all loss weights. 

\subsection{Appearance Editing}


With the bases predicted from the model, we can simply tune the value of the functions to support appearance editing, such as recoloring and photorealistic style transfer. When deployed with recent fast NeRF models, our method is able to achieve real-time interactive editing. We will further explain the editing details in the next section.

Since our basis functions are defined on the whole scene, they do not directly support local editing (e.g., edit a single object in the scene). To achieve that, we follow one recent work~\cite{DBLP:journals/corr/abs-2205-15585} that learns a 3D feature field from the semantic feature maps predicted from state-of-the-art image-based segmentation models (e.g., Lang-Seg~\cite{DBLP:conf/iclr/LiWBKR22}), and use the feature field to guide the editing. However, directly adding high dimensional semantic features to our model may decrease its efficiency, making it impractical to edit in real-time. As the objects captured from a scene are often limited to a small set (e.g., in an indoor scene, chairs, walls and floor are the most likely appeared objects),  the  semantic features extracted from the scene are often a limited subset of the whole feature space. Therefore, we apply PCA to compress the extracted features to a lower dimension (16 in our experiments) before feeding them to the network.~\footnote{Please find more details in terms of implementation, usage and results on semantic guided editing in the supplemental materials.} 

\section{Results}
\label{sec:results}

\subsection{Implementation details}
\paragraph{Training.} 
To support real-time appearance editing, we use Instant-NGP~\cite{Instant-NGP} as our backbone. \comment{Instant-NGP is a grid-based NeRF model that uses a feature hashing table to represent the 3D feature grid of the scene, which is used as the input position embeddings. It supports instant-level view synthesis in an interactive interface.}For the geometry learning stage, we keep all original configurations, except adding a per-point rgb loss introduced by Sun et al.~\cite{DVGO} to make the density field sparser and avoid floaters. For the segmentation learning stage, we fix the extracted palettes for the first 100 epochs. After that, we unleash the palettes and  remove $\mathcal{L}_{\textrm{weight}}$ to fine-tune the model. Moreover, since our smooth loss $\mathcal{L}_{\textrm{sm}}$ employs the diffuse color $\bm{c}_d$ to calculate the smoothing weight, we do not apply $\mathcal{L}_{\textrm{sm}}$ for the first 30 epoches to avoid unconverged $\bm{c}_d$ as input.

Our training and testing experiments are implemented using PyTorch~\cite{PyTorch}, and run on a single NVIDIA RTX 3090 GPU. For both stages, we train our model for 300-600 epochs (depending on the number of training images) using the Adam Optimizer~\cite{Adam} with learning rate set to $0.01$, which takes no more than 2 hours in total.

\paragraph{Datasets.} 
We conduct experiments on scenes from three sources: Lego, Ficus, Ship and Hotdog from the NeRF Blender dataset~\cite{NeRF}, Fern, Horns, Flower, Orchids from the the forward-facing LLFF dataset~\cite{LLFF} and Bonsai, Kitchen and Room from the 360-degree Mip-NeRF360 dataset~\cite{Mip-NeRF360}. 

\subsection{Comparisons}

\begin{figure}
\centering
\includegraphics[width=\linewidth]{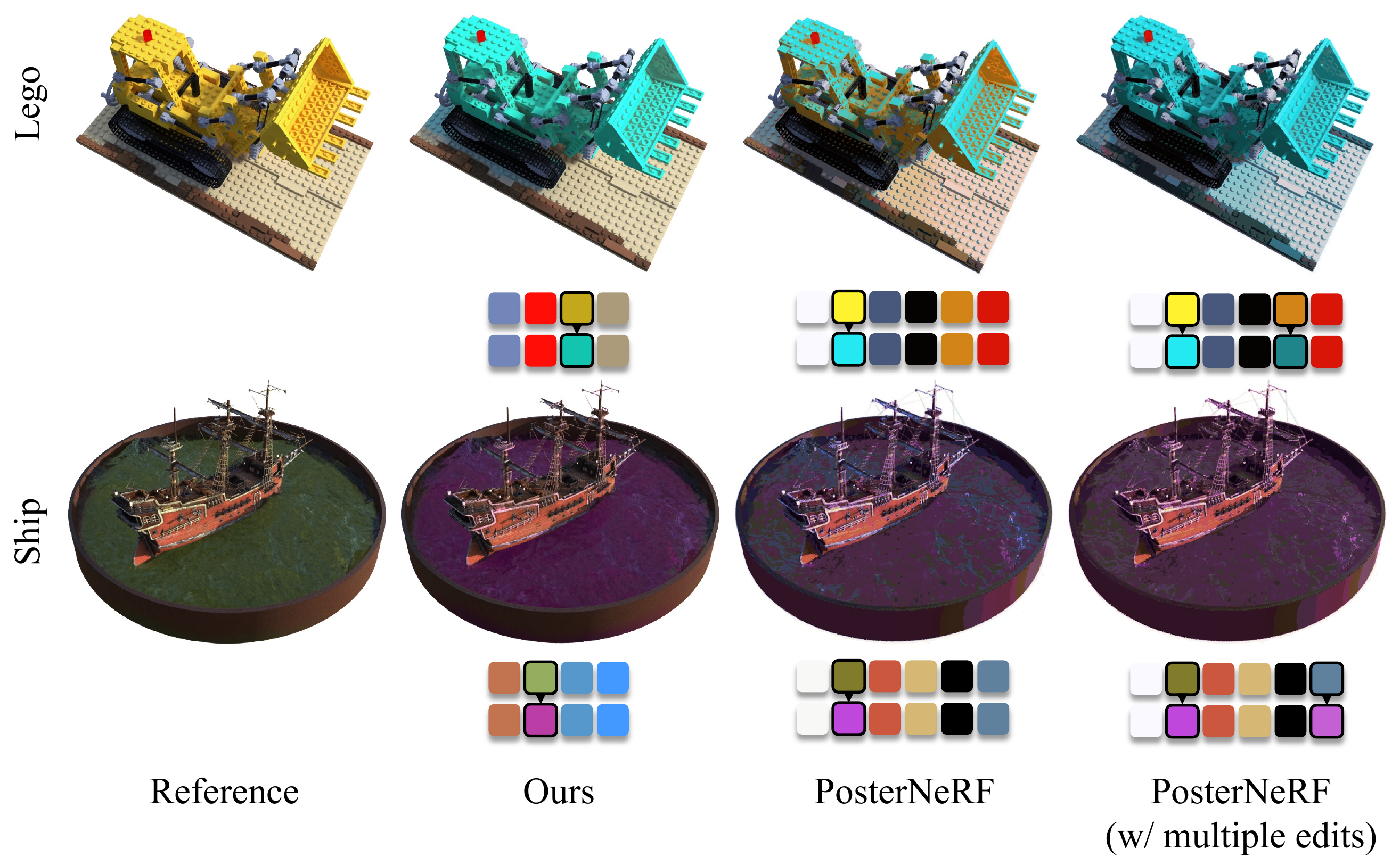}
\vspace{-10px}
\caption{\textbf{Qualitative comparison with PosterNeRF}~\cite{DBLP:journals/cgf/TojoU22}. Since PosterNeRF uses more palettes, we also show their results with edits on multiple palettes for fair comparison. Zoom in for details.}
\label{fig:posterNeRF}
\end{figure}

\begin{figure}
\centering
\includegraphics[width=\linewidth]{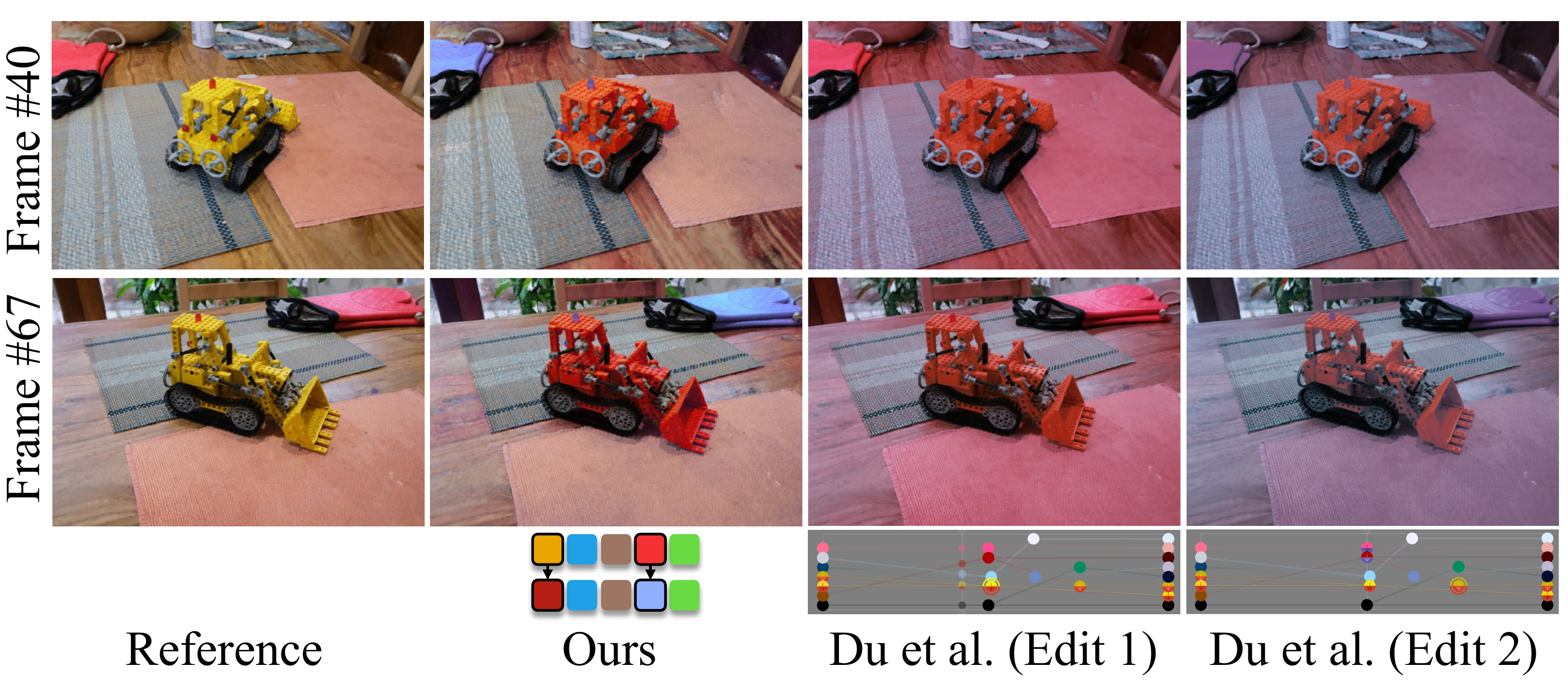}
\caption{\textbf{Qualitative comparison with video palette-based recoloring method}~\cite{Du:2021:VRS}. Unlike Du et al.~\cite{Du:2021:VRS} that either only modifies the lego color (Edit 1) or introduces view-inconsistency when also modifying the gloves (Edit 2), our method recolors the 3D scene consistently while maintaining photorealisim across views. }
\label{fig:video_recolor}
\vspace{-5px}
\end{figure}

\begin{figure*}
\centering
\includegraphics[width=0.99\textwidth]{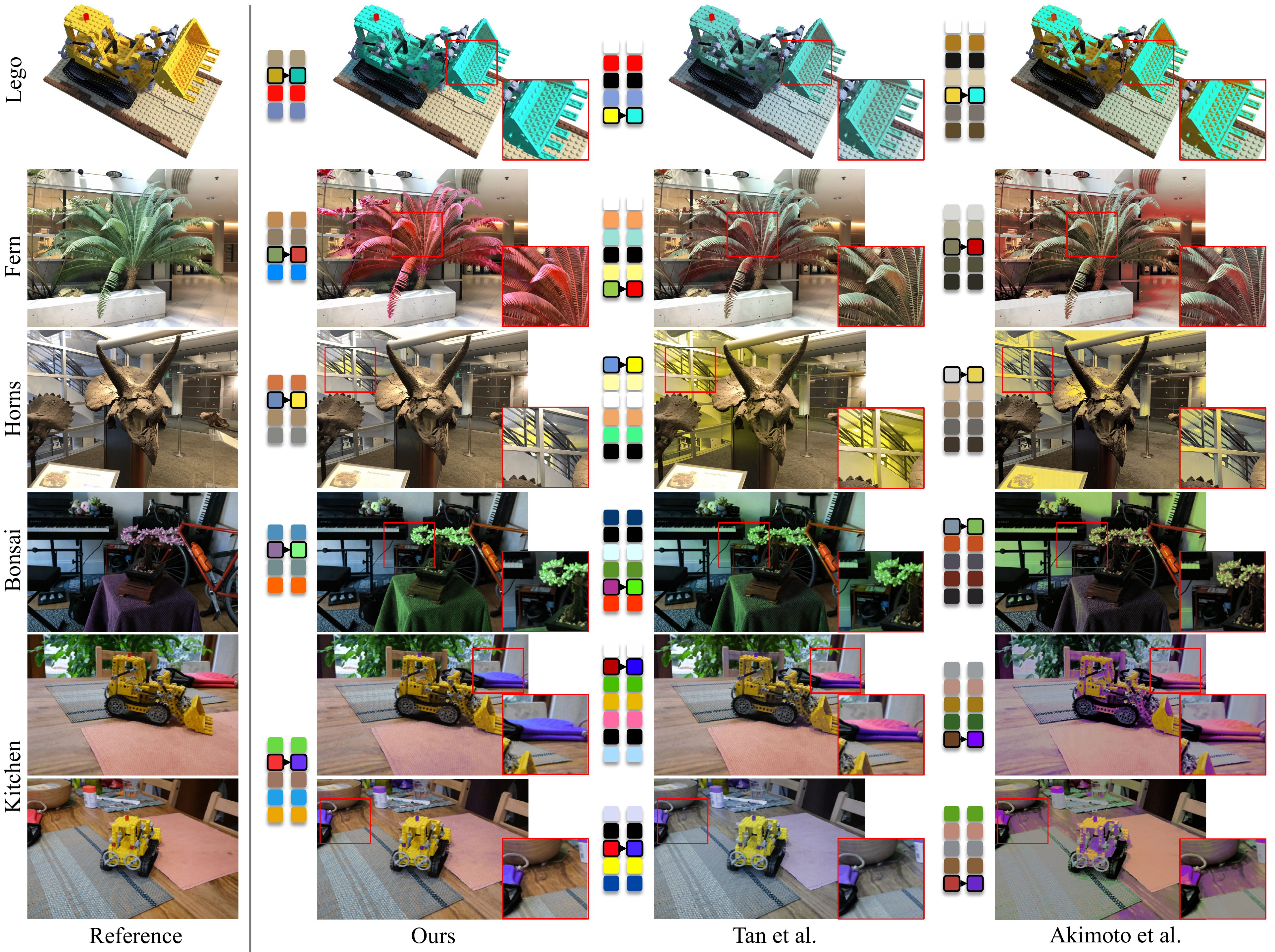}
\caption{\textbf{Qualitative comparison with image palette-based recoloring methods}. We compare our method with Tan et al.~\cite{RGBXY} and Akimoto et al.~\cite{FSCS}. For each recolored image, we also show the corresponding palettes editing on its left side, and a zoom-in view on its right side.}
\label{fig:recolor}
\vspace{-5px}
\end{figure*}

\begin{figure}[h]
\centering
\includegraphics[width=\linewidth]{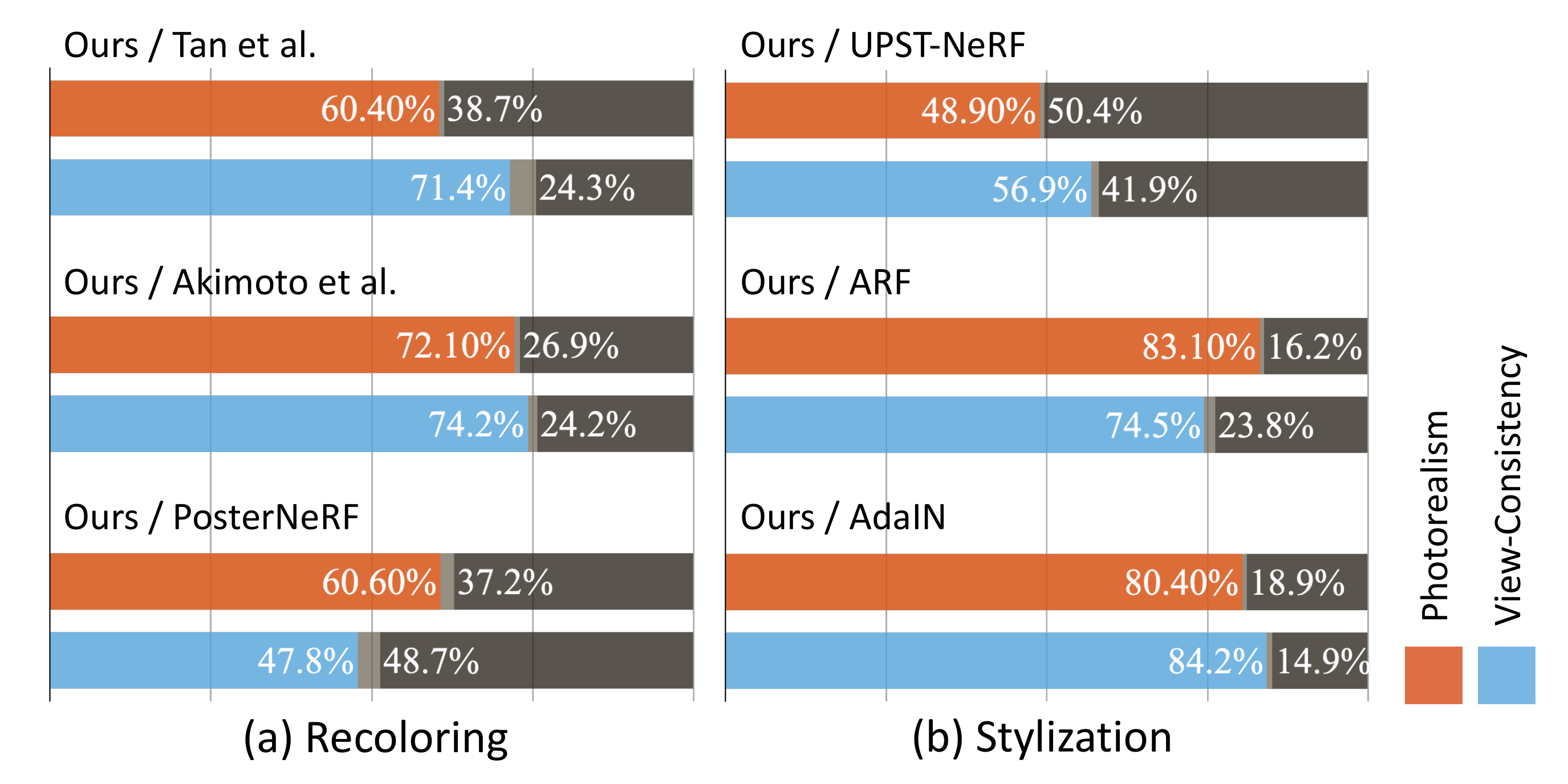}
\caption{\textbf{User study results.} For each comparison we demonstrate the percentage of users who prefer our method, users who prefer the baseline method, and users who do not lean towards anyone.}
\label{fig:user_study}
\vspace{-5px}
\end{figure} 

\begin{figure*}
\centering
\includegraphics[width=1.\textwidth]{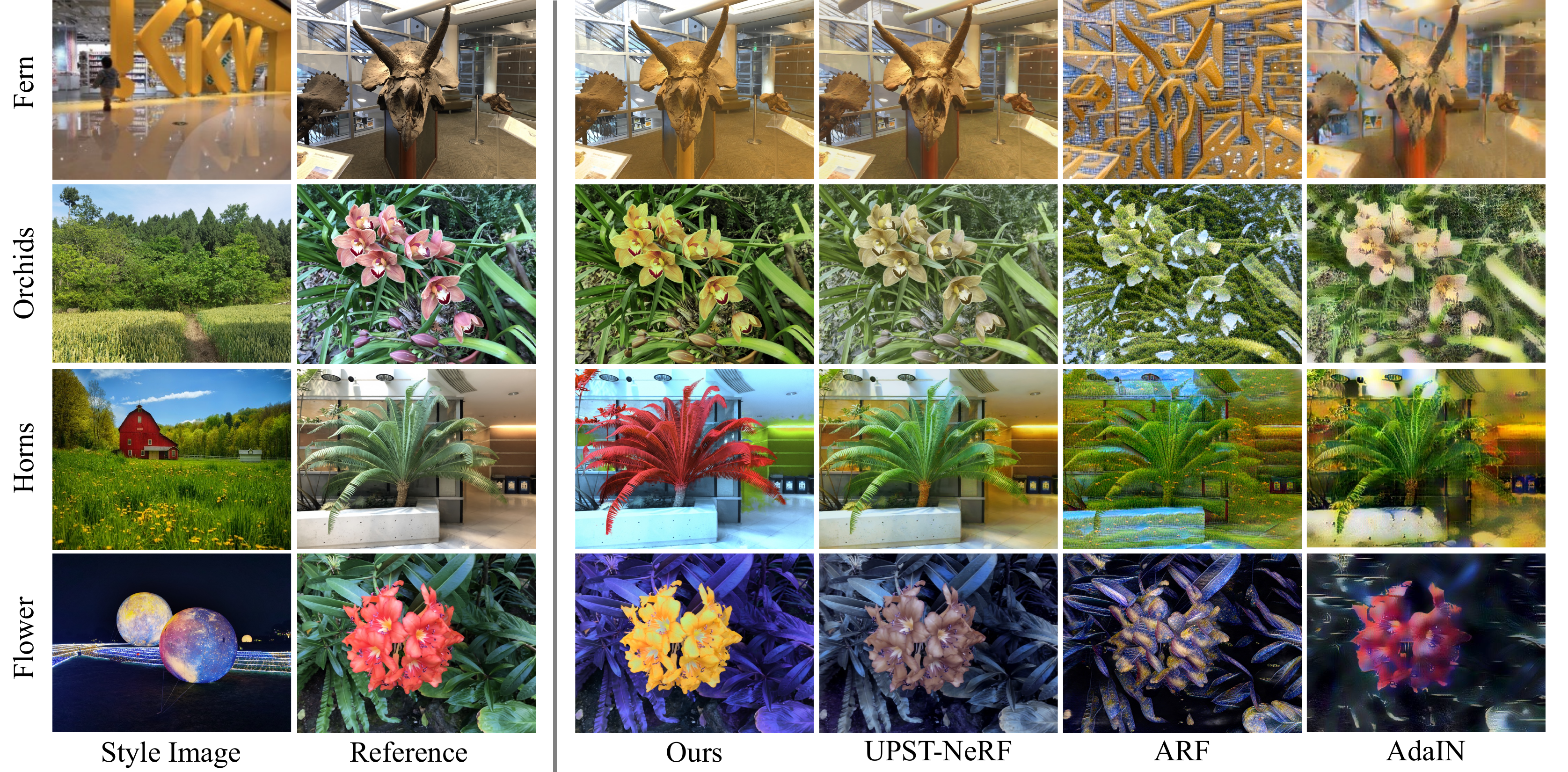}
\caption{\textbf{Qualitative comparison with baseline stylization methods}. Our method produces spatially and temporally consistent style transfer results that faithfully match the style examples while maintaining photorealism. }
\label{fig:stylization}
\end{figure*}

\paragraph{Recoloring.} 

We use the HSV color space for recoloring in all experiments. Given a group of modified color palettes $\bm{\mathcal{P}}'$, we calculate the difference between the original palettes $\bm{\mathcal{P}}$ and $\bm{\mathcal{P}}'$ in HSV, then directly apply the change to all points' soft color, i.e. $\bm{\mathcal{P}} + \bm{\delta}(\bm{x})$. 

We compare our model with the state-of-the-art NeRF-based recoloring method PosterNeRF~\cite{posterNeRF} on the Blender Dataset in Fig.~\ref{fig:posterNeRF}. Our model produces more photorealistic results with fewer artifacts, and requires fewer palette edits. 

We also compare our model with a state-of-the-art palette-based video recoloring model~\cite{du2021video} in Fig.~\ref{fig:video_recolor} and image recoloring models~\cite{RGBXY, FSCS} in Fig.~\ref{fig:recolor}. Since our model consists of color offsets and radiance functions that extend the palette color to basis, it requires fewer palettes to reconstruct the images than other methods, which improves usability and benefits the quality of recoloring results. Also, with the help of the scene geometry, our results are more plausible in 3D. Particularly, we show recoloring results on two different views in the Kitchen scene, where we set the same recoloring goal for each of them: changing the light on the excavator to purple. While the image-based methods failed to keep the other objects looking the same from the two views, our model is able to produce view-consistent results for the whole scene.

\paragraph{Photorealistic style transfer.} We can apply our decomposition model to achieve \emph{photorealistic} style transfer on the captured scenes. Given a style image, users may define a series of correspondences between the image pixels and 3D points, that our system uses to transform our palette-based bases to align the point colors to the pixel colors; more details in the supplementary. We compare our model with three state-of-the-art methods: UPST-NeRF~\cite{UPST-NeRF}, ARF~\cite{ARF} and image-based stylization with AdaIN~\cite{AdaIN} (Fig.~\ref{fig:stylization}). Our results are more photorealistic than ARF and AdaIN, and comparable with UPST-NeRF. However, UPST-NeRF takes much more time to train (tens of hours) than ours.

\paragraph{User study.} To better demonstrate the effectiveness of our model, we also conduct a user study on these two applications. The study was taken on the Amazon Mechanical Turk platform, where we dispatched 162 questions 30 times to the crowd (4860 in total), and received 4496 answers. In each question, the user watches images/videos generated from our model and another randomly selected baseline and was asked to decide which one is more view-consistent and/or photorealistic. Fig.~\ref{fig:user_study} summarizes our study's results. We outperform image-based recoloring baselines~\cite{RGBXY, FSCS}, and are considered more photorealistic than  posterNeRF~\cite{posterNeRF}. For the style transfer task, our model received better feedback than ARF~\cite{ARF} and AdaIN~\cite{AdaIN} and is rated better than UPST-NeRF~\cite{UPST-NeRF} for view-consistency. More details are in the supplementary.

\subsection{Ablations}

\begin{figure*}
\centering
\includegraphics[width=0.99\textwidth]{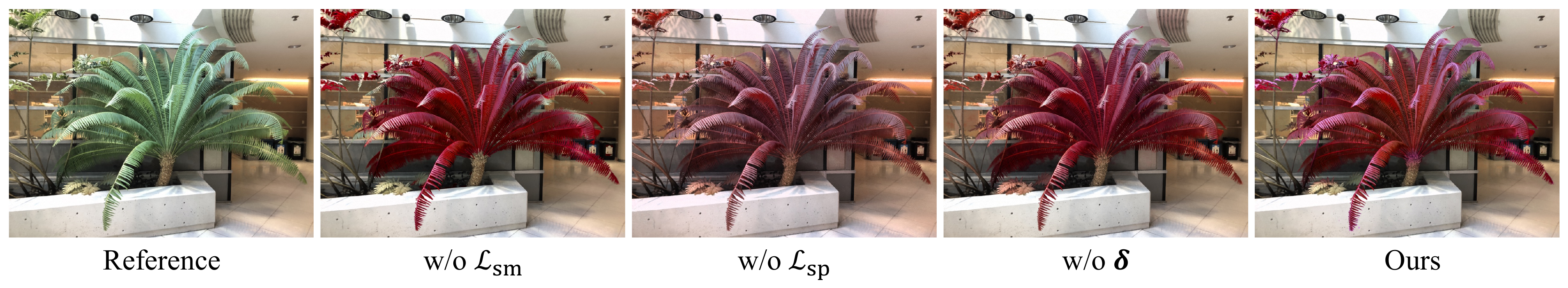}
\caption{\textbf{Qualitative ablation study results}. In this example we change the green palette (primarily representing the plants) to red color. }

\label{fig:qualitative_ablation}
\vspace{-5px}
\end{figure*}

We conduct two ablative evaluations on the LLFF dataset  to show the effectiveness of our model design. We first compare our model with the original vanilla Instant-NGP with two metrics: the Peak Signal-to-Noise Ratio (PSNR $\uparrow$), the Learned Perceptual Image Patch Similarity~\cite{LPIPS} (LPIPS$\downarrow$). Our model achieves the same PSNR score from the Instant-NGP (24.50), and has slightly higher LPIPS score (0.152 comparing to 0.145). This shows our model has little effect on the reconstruction quality. We then compare our model with three variants: Model without color offset (Ours w/o $\bm\delta$), model without the sparsity loss (Ours w/o $\mathcal{L}_\textbf{sp}$) and model without the smooth loss (Ours w/o $\mathcal{L}_\textbf{sm}$) both quantitatively and qualitatively. Quantitative results are shown in Tab.~\ref{tab:ablation}, where our full model achieves the best sparsity score among all models. Additionally, removing color offset will also decrease the quality of reconstruction in a perceptual way (LPIPS raises from 0.152 to 0.160). We also qualitatively shows comparisons in Fig.~\ref{fig:qualitative_ablation}, where removing smoothness loss leads to major artifacts in recoloring, and removing the sparsity loss results in color bleeding on unrelated areas (walls in the shown case).

\begin{table}
\begin{center}
\caption{\textbf{Quantitative ablation study results}. We compare our model with variants on two metrics: The sparsity error defined by Eq.~\ref{eq:sparse_loss} and the total variation of the weight images (TV) which measures their smoothness.}

\label{tab:ablation}
\newcolumntype{P}[1]{>{\centering\arraybackslash}p{#1}}
\vspace{-5px}
{\small
\begin{tabular}{P{53pt}P{38pt}P{38pt}}
Methods & Sparsity $\downarrow$ & TV  $\downarrow$  \\
\hline
Ours 
& \textbf{0.478} & 0.303  \\
Ours w/o $\bm{\delta}$ 
& 0.647 & 0.263  \\
Ours w/o $\mathcal{L}_{\textrm{sp}}$ 
& 1.643 &  \textbf{0.123}  \\
Ours w/o $\mathcal{L}_{\textrm{sm}}$  
& 0.554 & 0.549  \\
\hline

\end{tabular}}
\end{center}
\vspace{-10px}
\end{table}




\subsection{More Results}

\begin{figure}
\centering
\includegraphics[width=\linewidth]{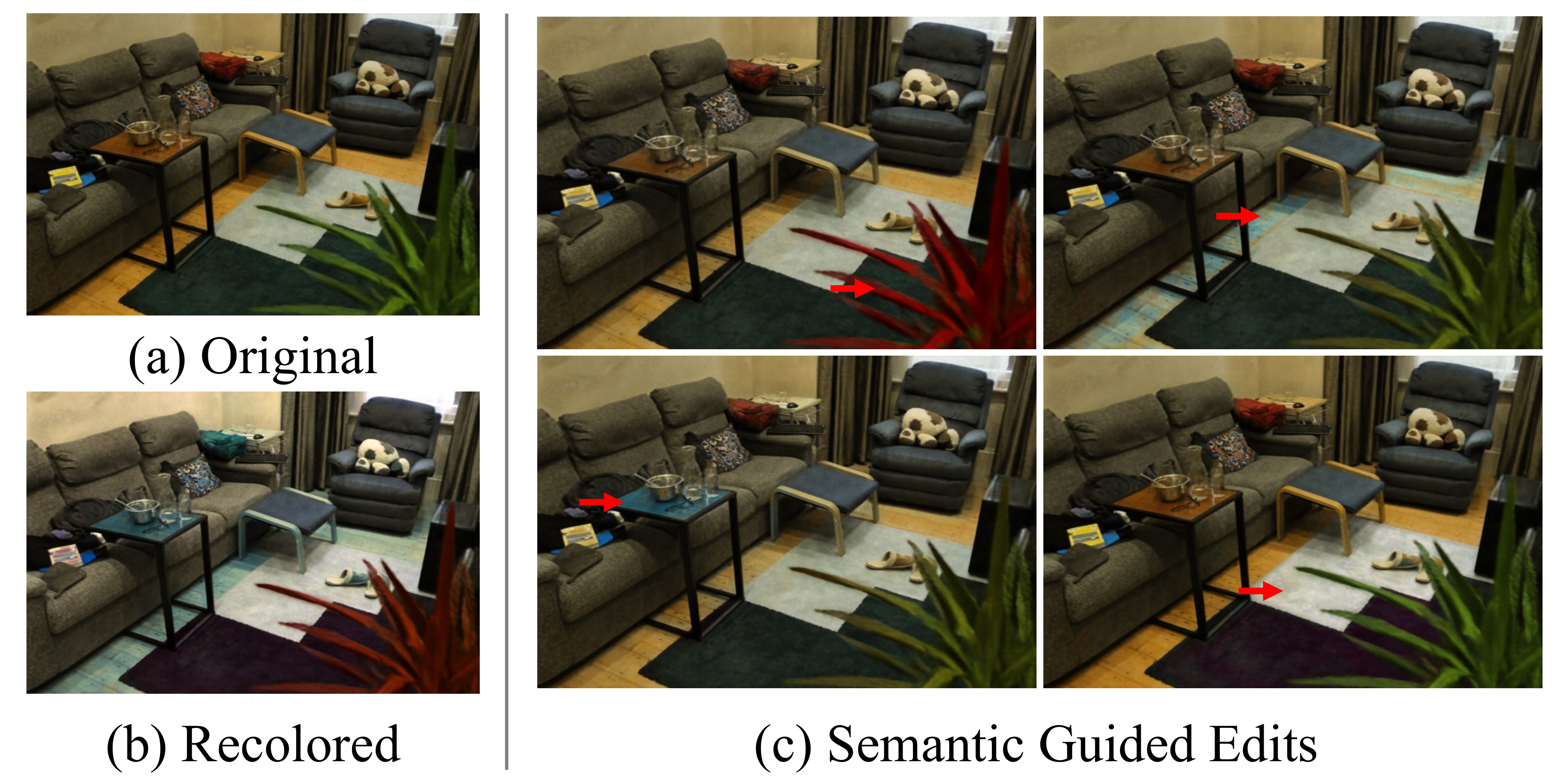}
\caption{\textbf{Semantic guided editing}. On the left we show the unedited image and recolored results w/o feature guidance. On the right side, we show four guided results where individual objects (plant, floor, table, carpet) are exclusively edited.}
\label{fig:CLIPediting}
\vspace{-15pt}
\end{figure}

In Fig.~\ref{fig:CLIPediting}, we show several results of feature-guided editing. Although our optimized segmentations are shared across the scene and do not directly support local editing, adding semantic features effectively resolves this issue.

\begin{figure}
\centering
\includegraphics[width=\linewidth]{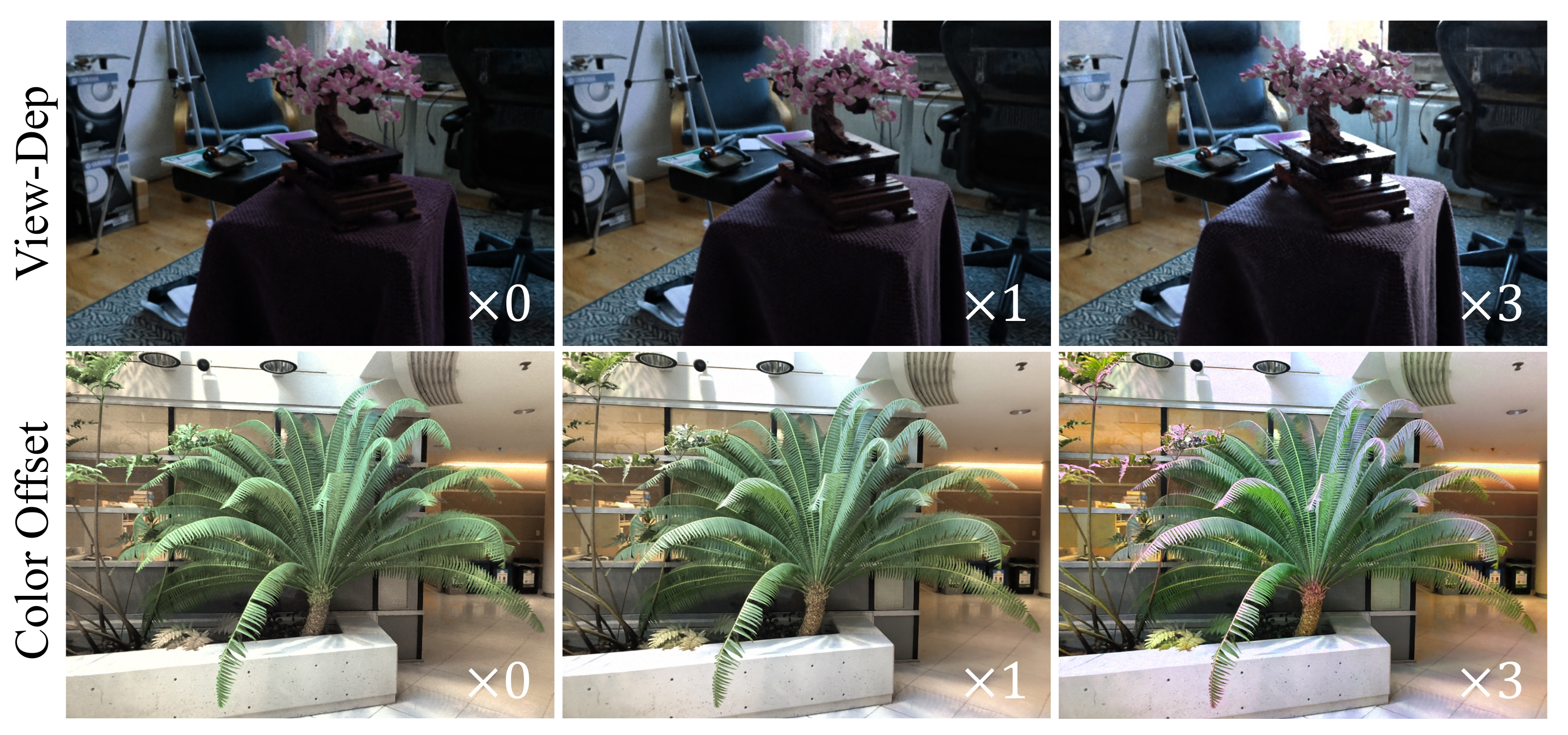}
\caption{\textbf{Additional appearance editing results.} \textit{First Row}: Rendered images with scaled view-dependent function; \textit{Second Row}: Rendered images with scaled color offset functions.}
\label{fig:application}
\end{figure}

In Fig.~\ref{fig:application}, we show two more edits supported by our decomposition results. By scaling the view-dependent color function $\bm{s}(\bm{x}, \bm{d})$ and the color offset functions $\bm{\delta}_i(\bm{x})$, we can modify the illumination conditions and object textures of the scene, while keeping the rendering photorealistic.
\section{Conclusion}
We have presented PaletteNeRF, a novel and efficient palette-based appearance editing framework for neural radiance fields. Our method significantly increases the practicality of palette-based appearance editing and enables intuitive and controllable interactive editing on real-world scenes. Our experiments illustrate the benefit of our approach for various editing tasks such as recoloring, photorealistic style transfer and illuminance changing. Future work may include frequency-based decomposition and editing of specular highlights and extending to dynamic NeRFs.
\paragraph{Acknowledgement}
This project was in part supported by Samsung, the Stanford Institute for Human-Centered AI (HAI), and a PECASE from the ARO. We also thank Jiaman Li for helping conducting our user study.

{\small
\bibliographystyle{ieee_fullname}
\bibliography{arxiv}
}

\end{document}